\documentclass{article}

\usepackage{PRIMEarxiv}

\usepackage[utf8]{inputenc} % allow utf-8 input
\usepackage[T1]{fontenc}    % use 8-bit T1 fonts
\usepackage{hyperref}       % hyperlinks
\usepackage{url}            % simple URL typesetting
\usepackage{booktabs}       % professional-quality tables
\usepackage{amsfonts}       % blackboard math symbols
\usepackage{nicefrac}       % compact symbols for 1/2, etc.
\usepackage{microtype}      % microtypography
\usepackage{lipsum}
\usepackage{fancyhdr}       % header
\usepackage{graphicx}       % graphics
\graphicspath{{media/}}     % organize your images and other figures under media/ folder

\usepackage{amsmath}
\usepackage{amsthm}
\theoremstyle{remark}
\newtheorem*{remark}{Remark}
\usepackage{amssymb}
\usepackage{algorithm}
\usepackage{algpseudocode}
\usepackage{multirow} 
\usepackage{makecell}
\usepackage{paralist}
\usepackage{comment}

\usepackage{stfloats}
\usepackage{enumitem}

\usepackage[dvipsnames]{xcolor}

\title{A Continuous-Time Consistency Model for 3D Point Cloud Generation}
\author{
  Sebastian Eilermann, René Heesch, Oliver Niggemann \\
  Institute for Artificial Intelligence  \\
  Helmut Schmidt University \\
  Hamburg\\
  \texttt{Surname.Lastname@hsu-hh.de} \\
}

\begin{document}
\definecolor{cvprblue}{rgb}{0.21,0.49,0.74}
\newtheorem{proposition}{Proposition}

\maketitle
\begin{abstract}
Fast and accurate 3D shape generation from point clouds is essential for applications in robotics, AR/VR, and digital content creation. We introduce \textbf{ConTiCoM-3D}, a continuous-time consistency model that synthesizes 3D shapes directly in point space, without discretized diffusion steps, pre-trained teacher models, or latent-space encodings. The method integrates a TrigFlow-inspired continuous noise schedule with a Chamfer Distance-based geometric loss, enabling stable training on high-dimensional point sets while avoiding expensive Jacobian-vector products. This design supports efficient one- to two-step inference with high geometric fidelity. In contrast to previous approaches that rely on iterative denoising or latent decoders, ConTiCoM-3D employs a time-conditioned neural network operating entirely in continuous time, thereby achieving fast generation. Experiments on the ShapeNet benchmark show that ConTiCoM-3D matches or outperforms state-of-the-art diffusion and latent consistency models in both quality and efficiency, establishing it as a practical framework for scalable 3D shape generation.
\end{abstract}

\section{Introduction}
\label{sec:intro}
Fast and accurate 3D shape generation from point clouds is essential for downstream tasks in robotics~\cite{wu2023fast}, autonomous driving~\cite{zeng2022lion}, medicine~\cite{peng2020convolutional,friedrich2023point}, and Augmented Reality (AR)/Virtual Reality (VR)~\cite{kim2021setvae}, where real-time inference is often required under hardware constraints.
In such settings, generative models must balance geometric fidelity with stringent runtime and memory limitations.
This rules out approaches that depend on hundreds of diffusion steps, expensive teacher–student distillation, or latent space compression, all of which are poorly suited for geometry-sensitive point cloud domains.

To address these challenges, a wide variety of 3D generative paradigms have been proposed, including variational autoencoders (VAEs)~\cite{kim2021setvae,petroll2023generative,hohmann2024design}, generative adversarial networks (GANs)~\cite{achlioptas2018learning,liu2018treegan}, flow-based models~\cite{yang2019pointflow,cai2020shapegf}, and increasingly popular diffusion models~\cite{ho2020denoising,song2020score,luo2021dpm,zhou2021pvd,friedrich2023point,mo2023dit3d}.
VAEs and GANs offer fast inference, but struggle with mode collapse and limited diversity.
Flow-based models achieve diversity and tractability, but require iterative sampling, which limits scalability~\cite{yang2019pointflow,kim2020softflow,klokov2020dpfnet,faubel2023towards,wu2023fast}.
Diffusion models produce high-quality samples but are slowed by multistep denoising~\cite{luo2021dpm,zhou2021pvd,mo2023dit3d}, while latent diffusion accelerates sampling at the cost of structural artifacts from encoder–decoder bottlenecks~\cite{zeng2022lion,du2024mlpcm}.
Point space approaches such as Point Straight Flow (PSF)~\cite{wu2023fast} reduce inference time but rely on hand-crafted heuristics that hinder generalization~\cite{faubel2023towards}.
Finally, consistency models~\cite{song2023consistency} promise one- to two-step generation, but existing training strategies do not scale to 3D: distillation-based CMs inherit the weaknesses of their diffusion teachers and fail without reliable perceptual metrics~\cite{song2023consistency,lu2024manicm}, Jacobian–vector product (JVP)-based supervision is prohibitively costly and unstable for high-dimensional point clouds~\cite{boffi2025build,song2023improved}, and latent space CMs~\cite{du2024mlpcm,dao2025improved} compress geometry through autoencoding, introducing artifacts.

In this work, we introduce \textbf{ConTiCoM-3D}, a \textbf{Con}tinuous-\textbf{Ti}me \textbf{Co}nsistency \textbf{M}odel for \textbf{3D} point cloud generation that operates directly in raw point space.
Unlike prior diffusion- or latent-based CMs, ConTiCoM-3D avoids autoencoding, handcrafted flows, teacher–student supervision, and unstable JVP-based losses.
Instead, it leverages a Chamfer Distance–based geometric loss with adaptive weighting, combined with a TrigFlow-inspired continuous noise schedule, to achieve stable training and efficient one- to two-step inference. The source code for the algorithm and to reproduce the evaluation results is available here \hyperlink{https://github.com/SEilermann/ConTiCoM_3D}{ConTiCoM\_3D}.

\noindent Our main contributions are:
\begin{compactitem}
\item \textbf{ConTiCoM-3D}, the first continuous-time consistency model for 3D point cloud generation that operates directly in raw point space, without relying on latent encodings, hand-crafted flows or teacher models.
\item A \textbf{geometry-aware adaptive loss} that combines Chamfer Distance with a learned time-dependent weighting function, providing directional consistency without costly or unstable JVPs.
\item Empirical evidence on ShapeNet, showing that ConTiCoM-3D achieves competitive one- to two-step sampling quality, surpassing diffusion and latent Consistency Models while enabling real-time inference.
\end{compactitem}

\section{Background}
\label{sec:background}
\begin{figure*}[tb]
  \centering
  \includegraphics[width=0.95\textwidth]{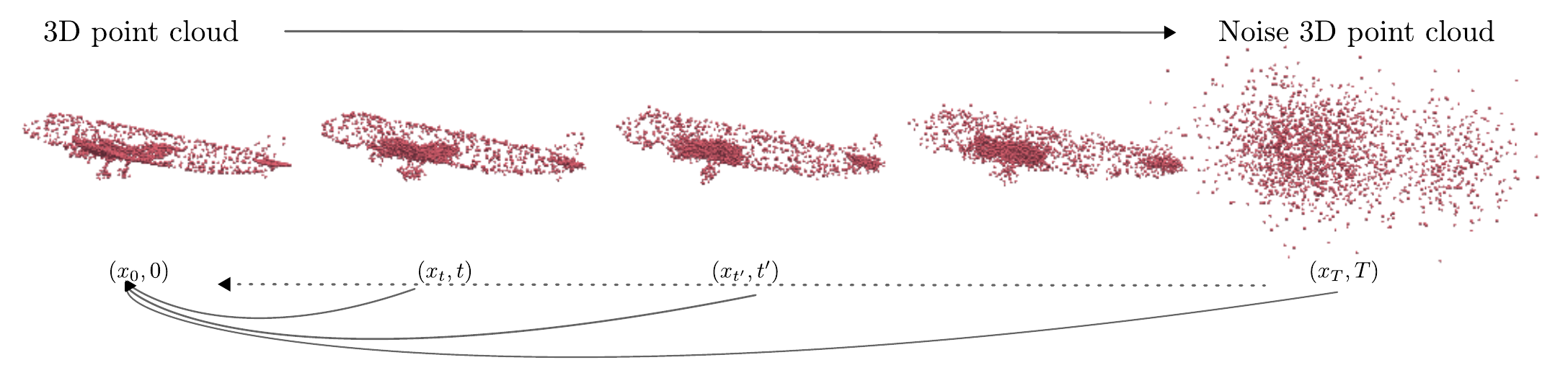}
  \caption{ConTiCoM-3D models a TrigFlow-inspired forward flow ODE (dashed line) that perturbs a clean point cloud $x_0$ into noisy samples $x_t, x_{t^{\prime}}$, and $x_T$. A time-conditioned model $f_\theta\left(x_t, t\right)$ learns to reverse this flow (dotted arrows), enabling single-step reconstruction and enforcing consistency along the continuous forward process without iterative denoising.}
  \label{fig:approach}
\end{figure*}
Our continuous-time 3D point cloud generation approach is based on diffusion models, Flow-Matching (FM) techniques, and Consistency Models (CMs). 
\subsection{Diffusion Models and Their Variants}
\label{sec:diffusion_models}
Recent advances in generative modeling have unified denoising diffusion probabilistic models (DDPMs)~\cite{ho2020denoising} and score-based methods~\cite{song2019generative, song2020score} within a continuous-time framework, where data is generated by solving an ODE that transforms noise into structure. Although DDPMs remain widely used, alternatives such as FM~\cite{lipman2022flow} and TrigFlow~\cite{lu2024simplifying} simplify training and sampling by directly learning velocity fields. These approaches improve stability, better align signal-to-noise ratios (SNRs), and are well-suited for consistency-based training.

Diffusion models~\cite{song2020score} reverse a standard forward noising process that perturbs a clean sample $x_0$ into $x_t = \alpha_t x_0 + \sigma_t z$, where $z \sim \mathcal{N}(0, I)$ is Gaussian noise. DDPMs~\cite{ho2020denoising} train a noise predictor $\epsilon_\theta$ and generate samples by solving the probability flow ODE ($\text{PF-ODE}$):
\begin{equation}
\dot{x}_t = \frac{d \log \alpha_t}{dt} x_t + \left( \frac{d \sigma_t}{dt} - \frac{d \log \alpha_t}{dt} \sigma_t \right) \epsilon_\theta(x_t, t).
\label{eq:ddpm_ode}
\end{equation}

FM~\cite{lipman2022flow} uses a linear schedule $\alpha_t = 1 - t$, $\sigma_t = t$, and trains a velocity field $v_\theta$ to match the reverse time velocity $v_t = \sigma_t' z - \alpha_t' x_0$, where $\alpha_t' = \frac{d\alpha_t}{dt}$ and $\sigma_t' = \frac{d\sigma_t}{dt}$. The model minimizes the following:
\begin{equation}
\mathbb{E}_{x_0, z, t} \left[ w(t)\left\| v_\theta(x_t, t) - (\sigma_t' z - \alpha_t' x_0) \right\|^2 \right],
\end{equation}
where $w(t)$ is a scalar time-dependent weighting function. Sampling then follows the simplified ODE $\frac{dx_t}{dt} = v_\theta(x_t, t)$.

TrigFlow~\cite{lu2024simplifying} adopts a spherical interpolation schedule with $\alpha_t = \cos(t)$ and $\sigma_t = \sin(t)$ for $t \in [0, \frac{\pi}{2}]$, yielding $x_t = \cos(t) x_0 + \sin(t) z$. It trains a time-conditioned velocity predictor $F_\theta$, scaled by a fixed noise level $\sigma_d > 0$, using:
\begin{equation}
\mathbb{E}_{x_0, z, t} \left[ w(t) \left\| \sigma_d F_\theta\left( \frac{x_t}{\sigma_d}, t \right) - (\cos(t) z - \sin(t) x_0) \right\|^2 \right],
\end{equation}

Samples are generated via:
\begin{equation}
\frac{dx_t}{dt} = \sigma_d F_\theta\left( \frac{x_t}{\sigma_d}, t \right).
\end{equation}
This formulation ensures consistent SNRs, simplifies time conditioning, and improves numerical stability, making it well suited for training consistency models. We adopt TrigFlow as the forward process for ConTiCoM-3D (see Sec.~\ref{sec:method}).

\subsection{Continuous-Time Consistency Models}
\label{sec:ctcm}
CMs~\cite{song2023consistency} directly learn a mapping $f_\theta(x_t, t)$ that predicts the clean sample $x_0$ from noisy input $x_t$ at time $t$. A common parameterization is as follows:
\begin{equation}
f_\theta(x_t, t) = c_{\text{skip}}(t) x_t + c_{\text{out}}(t) F_\theta(x_t, t),
\end{equation}
where $c_{\text{skip}}(t)$ and $c_{\text{out}}(t)$ are scalar functions that ensure $f_\theta(x_0, 0) = x_0$, and $F_\theta$ is a neural network.

\paragraph{Discrete-time vs. continuous-time.}
CMs can be trained in discrete or continuous time. Discrete-time models minimize discrepancies between adjacent steps with a stop-gradient teacher:
\begin{equation}
\mathbb{E}_{x_t, t} \!\left[ d\!\left(f_\theta(x_t, t), f_\theta^-(x_{t - \Delta t}, t - \Delta t)\right) \right],
\end{equation}
where $f_\theta^-$ is a stop-gradient copy and $d(\cdot,\cdot)$ is a distance metric such as squared error. However effective, this approach typically requires a pre-trained diffusion teacher, increasing cost and constraining performance.

Continuous-time CMs~\cite{lu2024simplifying} avoid discretization by taking $\Delta t \to 0$, leading to:
\begin{equation}
\mathcal{L}_{\text{cont.CM}} = \mathbb{E}_{x_t, t}\!\left[w(t)\,\big\langle f_\theta(x_t, t), \tfrac{d}{dt} f_\theta(x_t, t)\big\rangle\right].
\end{equation}
Although elegant in theory, this requires JVPs to compute $\tfrac{d}{dt}f_\theta$~\cite{boffi2025build,song2023improved}, which scale poorly in high-dimensional 3D point clouds.

\paragraph{Practical workarounds.}
Alternative strategies such as distillation~\cite{song2023consistency,lu2024manicm} introduce the dependence on pre-trained diffusion teachers, which both increase training cost and limit performance at the teacher’s quality level. Latent consistency models~\cite{du2024mlpcm,dao2025improved} sidestep these issues through compression, but suffer from encoder–decoder bottleneck artifacts. As a result, directly applying existing CM objectives to point clouds leads to instability and degraded geometry.

\paragraph{TrigFlow parameterization.}
Under the TrigFlow path, the continuous-time CM has a simple closed form:
\begin{equation}
f_\theta(x_t, t) = \cos(t) x_t - \sin(t)\,\sigma_d F_\theta\!\left(\tfrac{x_t}{\sigma_d}, t \right),
\label{eq:ctcm}
\end{equation}
where $\sigma_d > 0$ is the data scale. This form avoids ODE solvers and supports efficient training. However, existing objectives still rely on JVPs or distillation, leaving open the question of how to train CMs for 3D efficiently and robustly.

\paragraph{Motivation for ConTiCoM-3D.}
These limitations motivate ConTiCoM-3D: a teacher-free, JVP-free, latent-free continuous-time CM tailored to 3D point clouds. Unlike DDPMs and FM models that require many inference steps, ConTiCoM-3D enables fast single-step sampling. In contrast to JVP-based continuous-time consistency models~\cite{boffi2025build}, we avoid unstable Jacobian–vector products, which are prohibitively expensive for unordered 3D point clouds. Our approach also avoids teacher–student distillation~\cite{song2023consistency}, which relies on strong perceptual metrics that are not available in point space, and sidesteps latent compression~\cite{du2024mlpcm}, which introduces shape artifacts. Together, these design choices allow our model to operate robustly and efficiently in the raw point cloud domain without teachers, JVPs, or latent bottlenecks, while preserving geometric fidelity through a lightweight Chamfer reconstruction term (see Sec.~\ref{sec:method}).

\section{Related Work}
\label{sec:sota}
\paragraph{Adversarial and Autoencoding Approaches.}
The early 3D generative models adopted adversarial frameworks such as r-GAN~\cite{achlioptas2018learning}, TreeGAN~\cite{liu2018treegan}, SP-GAN~\cite{li2021spgan}, and PDGN~\cite{hui2020pdgn}.  While pioneering, these methods suffer from training instability, mode collapse, and under-constrained geometry. PDGN alleviates some of these issues by progressively generating multi-resolution point clouds with a shape-preserving adversarial loss, but it remains constrained by adversarial optimization. Variational autoencoders (VAEs)~\cite{kim2021setvae,petroll2023generative,hohmann2024design} and graph-based extensions such as GCA~\cite{zhang2021learning} improved stability but compress geometry through latent bottlenecks, limiting the fidelity for fine-grained shape reconstruction.

\paragraph{Adversarial and Autoencoding Approaches.}
The early 3D generative models adopted adversarial frameworks such as r-GAN~\cite{achlioptas2018learning}, TreeGAN~\cite{liu2018treegan}, and SP-GAN~\cite{li2021spgan}.  
While pioneering, these methods suffer from training instability, mode collapse, and under-constrained geometry.  
Variational autoencoders (VAEs)~\cite{kim2021setvae,petroll2023generative,hohmann2024design} improved stability but compress geometry through latent bottlenecks, limiting the fidelity for fine-grained shape reconstruction.

\paragraph{Flow-Based Models.}
Flow-based methods such as PointFlow~\cite{yang2019pointflow} and ShapeGF~\cite{cai2020shapegf} model continuous distributions over point sets.  
They offer exact likelihoods and diverse sampling, but require expensive ODE integration at inference.  
Later refinements like SoftFlow~\cite{kim2020softflow}, DPF-Net~\cite{klokov2020dpfnet}, and PSF~\cite{wu2023fast} reduce inference time, but often rely on handcrafted priors or dynamics that limit generalization.

\paragraph{Diffusion Models.}
Diffusion models~\cite{ho2020denoising,song2020score} and their 3D extensions~\cite{luo2021dpm,zhou2021pvd,mo2023dit3d,friedrich2023point} achieve state-of-the-art fidelity by progressively denoising from Gaussian noise.  
However, they require hundreds of steps, preventing real-time deployment.  
Latent diffusion variants, such as LION~\cite{zeng2022lion}, MLPCM~\cite{du2024mlpcm}, and MeshDiffusion~\cite{liu2023meshdiffusion}, accelerate inference but introduce discretization artifacts from encoder–decoder compression.

\paragraph{Consistency Models.}
CMs~\cite{song2023consistency} promise one- to two-step generation by enforcing forward process alignment in continuous flows.  
Distillation-based CMs~\cite{song2023consistency,lu2024manicm} inherit the limitations of their diffusion teachers and are weak in 3D domains without reliable perceptual metrics.  
Continuous-time CMs~\cite{lu2024simplifying,song2023improved,boffi2025build} replace teachers with JVP supervision, which is costly and unstable for high-dimensional point sets.  
Latent space CMs~\cite{du2024mlpcm,dao2025improved} reduce training cost but compress geometry, introducing bottleneck artifacts.  
Recent extensions include multistep CMs~\cite{heek2024multistep}, inverse-flow consistency~\cite{zhang2025inverse}, and one-step acceleration methods such as SANA~\cite{chen2025sana} or alignment-based strategies~\cite{sabour2025align}.  
While effective in images or robotics~\cite{lu2024manicm}, these designs either trade efficiency for fidelity or depend on surrogate objectives.  
In contrast, ConTiCoM-3D is the first CM to operate \emph{directly in raw point space} without teachers, JVPs, or latent bottlenecks, using only analytic flow supervision and a Chamfer-based reconstruction loss.

%\paragraph{Summary.}
%Overall, the landscape of 3D generative modeling reveals a persistent trade-off.  
%Adversarial and autoencoding methods are efficient but limited in fidelity and diversity.  
%Flow- and diffusion-based models provide strong geometric quality but are hindered by the need for many iterative steps.
%Latent and multistep consistency models mitigate cost but either introduce artifacts or add algorithmic complexity.  
%These challenges motivate the design of \textbf{ConTiCoM-3D}, a continuous-time CM that is teacher-free, JVP-free and latent-free, enabling robust single-step point cloud generation directly in raw point space.
\paragraph{Summary.}  
Overall, the landscape of 3D generative modeling reveals a persistent trade-off. Adversarial and autoencoding methods~\cite{achlioptas2018learning,liu2018treegan,li2021spgan,kim2021setvae,petroll2023generative,hohmann2024design} are efficient but limited in fidelity and diversity, while flow- and diffusion-based models~\cite{yang2019pointflow,cai2020shapegf,kim2020softflow,klokov2020dpfnet,faubel2023towards,ho2020denoising,song2020score,luo2021dpm,zhou2021pvd,mo2023dit3d,friedrich2023point} provide strong geometric quality but require many iterative steps. Latent and multistep consistency models~\cite{zeng2022lion,du2024mlpcm,dao2025improved,heek2024multistep,chen2025sana} mitigate cost but either introduce artifacts or add algorithmic complexity. Moreover, existing CM variants do not transfer cleanly to unordered 3D point clouds. Distillation-based CMs~\cite{song2023consistency,lu2024manicm} rely on perceptual similarity metrics that are effective in image space but unavailable in point space, leading to weak supervision signals. Continuous-time CMs with JVP supervision~\cite{song2023improved,boffi2025build} are computationally prohibitive and often unstable when applied to high-dimensional shapes. Latent-space CMs~\cite{du2024mlpcm,dao2025improved} mitigate these costs but compress geometry through encoder--decoder bottlenecks, introducing structural distortions. These limitations motivate the design of \textbf{ConTiCoM-3D}, a continuous-time CM that is teacher-free, JVP-free, and latent-free, combining analytic FM with Chamfer-based supervision to enable robust one- to two-step point cloud generation directly in raw point space.  

\begin{algorithm*}[t]
\caption{ConTiCoM-3D: JVP-free training with flow-matching and Chamfer reconstruction}
\label{alg:ctcm_train_fm}
\begin{algorithmic}[1]
\Require Dataset $\mathcal{X}$; velocity net $F_\theta$; symmetric squared Chamfer $\mathrm{CD}$; data scale $\sigma_d$; $T_{\max}=\pi/2$
\For{each epoch}
  \For{each batch $x_0 \sim \mathcal{X}$}
    \State Sample $t \sim \mathcal{U}(0, T_{\max})$, $z \sim \mathcal{N}(0, \sigma_d^2 I)$
    \State Construct noisy sample $x_t=\cos t\,x_0+\sin t\,z$ \hfill (Eq.~\ref{eq:trigflow})
    \State Compute velocity $v_\theta=F_\theta(x_t/\sigma_d, t)$
    \State Predictor $f_\theta(x_t,t)=\cos t\,x_t-\sin t\,\sigma_d\,v_\theta$ \hfill (Eq.~\ref{eq:ctcm_form})
    \State \textbf{FM loss:} $\mathcal{L}_{\mathrm{FM}}=\|\,\sigma_d\,v_\theta-(\cos t\,z-\sin t\,x_0)\|_2^2$ \hfill (Eq.~\ref{eq:fm})
    \State \textbf{Chamfer loss:} $\mathcal{L}_{\mathrm{CD}}=\mathrm{CD}(f_\theta(x_t,t),x_0)$ \hfill (Eq.~\ref{eq:cd})
    \State \textbf{Total:} $\mathcal{L}_{\text{total}}=\mathcal{L}_{\mathrm{FM}}+\lambda_{\mathrm{CD}}\,\mathcal{L}_{\mathrm{CD}}$ \hfill (Eq.~\ref{eq:total_loss})
    \State Update $\theta$ by SGD/Adam
  \EndFor
\EndFor
\end{algorithmic}
\end{algorithm*}

\section{Method}
\label{sec:method}
We propose \textbf{ConTiCoM-3D}, the first continuous-time consistency model that operates directly in raw point space and achieves single-step point cloud generation.  
Unlike previous approaches, ConTiCoM-3D is \emph{teacher-free, JVP-free, and latent-free}: it avoids diffusion teachers, costly Jacobian supervision, and lossy latent compression.  
Our method relies on two complementary objectives: an analytic FM regression target and a lightweight Chamfer reconstruction term.  
Together, these provide geometry-aware, permutation-invariant supervision that is both stable and efficient, enabling real-time inference with one to four sampling steps.

The complete training loop is summarized in Algorithm~\ref{alg:ctcm_train_fm}, where each step corresponds to its formal definition in the text.

\subsection{Problem Definition}
Let $\mathcal{X} = \{X_i\}_{i=1}^N$ be a dataset of different 3D point clouds, with each $X_i\in\mathbb{R}^{M\times 3}$ containing unordered points.  
Following TrigFlow~\cite{lu2024simplifying} (see Sec.~\ref{sec:diffusion_models}), the forward process is as follows:
\begin{equation}
x_t = \cos t\,x_0 + \sin t\,z,\qquad z \sim \mathcal{N}(0,\sigma_d^2 I),
\label{eq:trigflow}
\end{equation}
with $t \in [0, T_{\max}]$ and $T_{\max}=\tfrac{\pi}{2}$ (see Alg.~\ref{alg:ctcm_train_fm}, l.~3–4).  
The closed-form predictor is parameterized as

The predictor admits a closed-form expression derived by substituting the analytic TrigFlow velocity into the CT-CM parameterization of Eq.~\ref{eq:ctcm}:  
\begin{equation}
f_\theta(x_t,t)=\cos t\,x_t-\sin t\,\sigma_d\,F_\theta(x_t/\sigma_d,t).
\label{eq:ctcm_form}
\end{equation}
This form enforces $f_\theta(x,0)=x$ and realizes the continuous-time consistency formulation described in Sec.~\ref{sec:ctcm} (see Alg.~\ref{alg:ctcm_train_fm}, l.~6). Here, $\mathrm{CD}$ denotes the symmetric squared Chamfer distance, normalized by $M$ points.

\subsection{Model Training}
\label{sec:training}
\paragraph{Motivation.}  
Temporal derivative supervision in continuous-time CMs usually requires JVPs~\cite{boffi2025build}, which are unstable and computationally heavy in high-dimensional geometry.  
Distillation-based supervision~\cite{song2023consistency,jiang2025consistency} depends on diffusion teachers and fails in 3D due to the absence of robust perceptual metrics.  
We therefore adopt a \emph{single-time, JVP-free} supervision that combines an analytic flow target with a lightweight Chamfer term.  
This design is efficient, unbiased, and stable for unordered point clouds.

\paragraph{Analytic flow matching.}  
The TrigFlow forward process admits an analytic velocity
\begin{equation}
\frac{dx_t}{dt} = -\sin t\,x_0+\cos t\,z,
\label{eq:forward_trigflow}
\end{equation}
yielding the regression objective
\begin{equation}
\mathcal{L}_{\mathrm{FM}}
=\big\|\,\sigma_d\,F_\theta(x_t/\sigma_d,t)-(\cos t\,z-\sin t\,x_0)\,\big\|_2^2.
\label{eq:fm}
\end{equation}
corresponding to line~7 in Alg.~\ref{alg:ctcm_train_fm} and related to Flow Matching~\cite{lipman2022flow}. Unlike secant-based consistency losses~\cite{chen2025convergence}, this is unbiased, variance-reduced, and avoids higher-order derivatives.

\begin{proposition}[Closed-form recovery]
\label{prop:exact_recovery}
If $F_\theta$ matches the analytic velocity, then $f_\theta(x_t,t)=x_0$ for all $t\in[0,T_{\max}]$.
\end{proposition}
This implies that perfect velocity regression suffices for exact reconstruction, without the need for iterative refinement (see proof in Appendix~\ref{sec:proposition_proof}).

\paragraph{Chamfer reconstruction.}  
To encourage geometric fidelity, we add a permutation-invariant Chamfer loss~\cite{fan2017point}:
\begin{equation}
\mathcal{L}_{\mathrm{CD}}=\mathrm{CD}(f_\theta(x_t,t),x_0).
\label{eq:cd}
\end{equation}
This corresponds to line~8 in Alg.~\ref{alg:ctcm_train_fm}.

\paragraph{Total objective.}  
The final training loss is
\begin{equation}
\mathcal{L}_{\text{total}}=\mathcal{L}_{\mathrm{FM}}+\lambda_{\mathrm{CD}}\,\mathcal{L}_{\mathrm{CD}}.
\label{eq:total_loss}
\end{equation}
which corresponds to line~9 in Alg.~\ref{alg:ctcm_train_fm}.  
Unlike previous CMs~\cite{song2023consistency,du2024mlpcm}, we train without an EMA teacher.

\subsection{Sampling and Inference}
\label{sec:sampling}
\paragraph{Single-step generation.}  
Sampling requires only one evaluation at $T=T_{\max}$:
\begin{equation}
\hat{x}_0=f_\theta(x_T,T),\qquad x_T\sim\mathcal{N}(0,\sigma_d^2 I).
\label{eq:singlestep}
\end{equation}
This enables practical single-step point cloud generation, unlike diffusion models that require hundreds of denoising steps.

\paragraph{Few-step refinement.}  
For optional higher fidelity, we define a total number of inference steps $S \in \mathbb{N}$ and partition $[0, T_{\max}]$ into $S$ uniform time intervals of size $\Delta = T_{\max} / S$.
We integrate the TrigFlow $\text{PF-ODE}$ in reverse time using explicit Euler steps:
\begin{equation}
x_{t - \Delta} = x_t - \Delta\, \sigma_d\, F_\theta(x_t/\sigma_d, t),
\label{eq:pfode_sampler}
\end{equation}
with $t = T_{\max}, T_{\max} - \Delta, \dots, \Delta$.
A Heun proposal~\cite{lu2024simplifying} may optionally be used for local error control, while accepted states follow the Euler update.
Empirically, small values of $S$ already suffice for high-fidelity generation with minimal cost~\cite{lu2024simplifying,song2023consistency,chen2025sana}.

\section{Experiments}
\label{sec:experiments}
We evaluate ConTiCoM-3D on standard 3D point cloud generation benchmarks in both single-class and multi-class settings. The experimental design focuses on three aspects: the fidelity and diversity of generated point clouds, the efficiency of inference in terms of sampling speed, and the robustness of the method, as demonstrated through ablation studies.  

\subsection{Experimental Setup}
\label{sec:setup}
Following prior work~\cite{yang2019pointflow,zeng2022lion,du2024mlpcm}, we evaluate the generative performance on ShapeNet~\cite{chang2015shapenet} for single-class generation (\textit{airplane, chair, car}) and on ShapeNet-vol~\cite{peng2021shape} for multi-class generation across 13 categories. All shapes are uniformly resampled to 2{,}048 points.  
The evaluation is based on \textbf{1-Nearest Neighbor Accuracy (1-NNA)} computed under Chamfer Distance (CD) and Earth Mover’s Distance (EMD)~\cite{yang2019pointflow}. Lower scores indicate a better balance between fidelity and diversity.  

Our model builds on a Point-Voxel CNN (PVCNN)~\cite{liu2019pvcnn} U-Net backbone, augmented with PointNet++~\cite{qi2017pointnet++} set abstraction and feature propagation modules. Time conditioning is introduced at the bottleneck via sinusoidal embeddings. Training is performed for 4 days on two NVIDIA L40S GPUs using the Adam optimizer~\cite{kingma2014adam} with a fixed learning rate of $1 \times 10^{-4}$. More architectural and training details are provided in the Appendix~\ref{sec:architecture}.

\subsection{Single-Class 3D Point Cloud Generation}
\label{sec:single_class_gen}

\begin{figure*}[htb]
  \centering
  \includegraphics[width=0.95\textwidth]{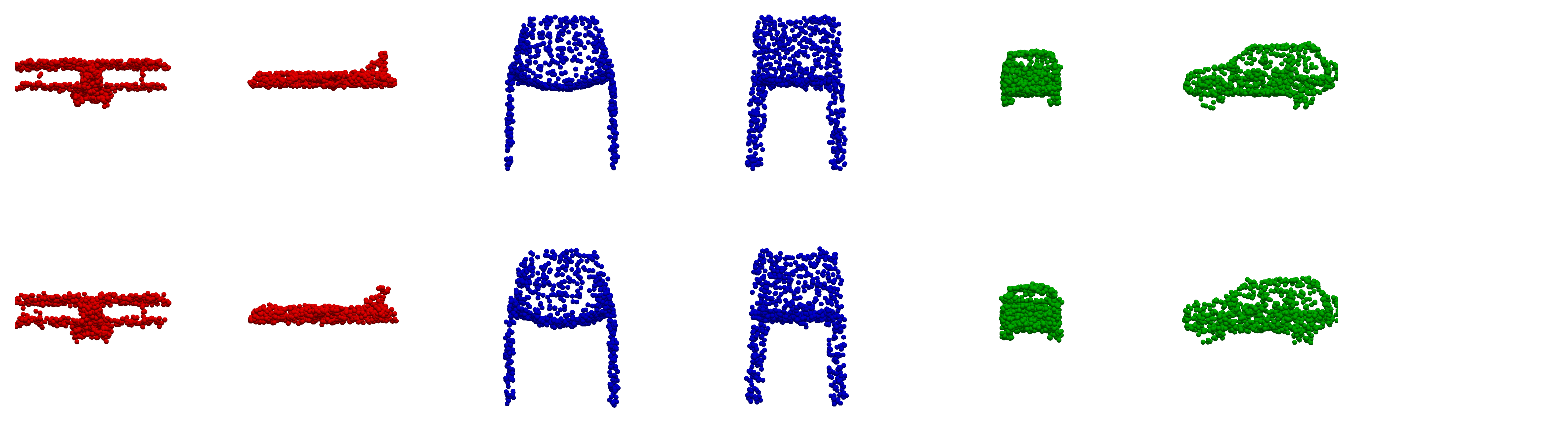}
  \caption{Single-class generation results (\textit{airplane, chair, car}). Top row: samples generated by ConTiCoM-3D with $S=1$ step. Bottom row: samples from MLPCM~\cite{du2024mlpcm}. ConTiCoM-3D produces sharper structures and more consistent global geometry while maintaining diversity, whereas MLPCM exhibits compression artifacts from latent bottlenecks.}
    \label{fig:single_gnerated}
\end{figure*}

We first consider the setting where the models are trained separately for each category (\textit{airplane, chair, car}), following the protocol of PointFlow~\cite{yang2019pointflow}. A visualization of results is given in Figure~\ref{fig:single_gnerated}.

\noindent\textbf{Quantitative Results.}  
Table~\ref{tab:sing_gen} and Table~\ref{tab:single_class_individuell} shows that ConTiCoM-3D consistently outperforms GAN- and VAE-based baselines and is competitive with recent diffusion and consistency models. With only two inference steps, our method achieves the lowest EMD scores and competitive CD performance across all three categories, while only requiring two evaluations at inference. Compared to latent approaches such as LION~\cite{zeng2022lion} and MLPCM~\cite{du2024mlpcm}, ConTiCoM-3D operates directly in the raw point space, avoiding encoder–decoder artifacts and enabling temporally stable and spatially precise generation. Moreover, its runtime is significantly faster than multistep diffusion models, offering a strong balance between speed and fidelity.
\\
\\

\begin{table}[htb]
\centering
\caption{Single-class generation results on \textit{airplane, chair, car} categories on ShapeNet dataset from PointFlow~\cite{yang2019pointflow}. We use 1-NNA$\downarrow$ as generation metric for evaluation. Training and test data normalized globally into $[-1, 1]$. We observe that our model achieves a competitive balance across speed and accuracy.}
\setlength{\tabcolsep}{3pt} % tighter columns
\renewcommand{\arraystretch}{1.05} % a touch more row height (readability)
\begin{tabular}{lrrrrrr}
\toprule
\multirow{2}{*}{\textbf{Method}} &
\multicolumn{2}{c}{\textbf{Airplane}} &
\multicolumn{2}{c}{\textbf{Chair}} &
\multicolumn{2}{c}{\textbf{Car}} \\
& \textbf{CD} & \textbf{EMD} &
\textbf{CD} & \textbf{EMD} &
\textbf{CD} & \textbf{EMD} \\
\midrule
r-GAN~\cite{achlioptas2018learning} & 98.40 & 96.79 & 83.69 & 99.70 & 94.46 & 99.01 \\
1-GAN(CD)~\cite{achlioptas2018learning} & 87.30 & 93.95 & 68.58 & 83.84 & 66.49 & 88.78 \\
1-GAN(EMD)~\cite{achlioptas2018learning} & 89.49 & 76.91 & 71.90 & 64.65 & 71.16 & 66.19 \\
PointFlow~\cite{yang2019pointflow} & 75.68 & 70.74 & 62.84 & 60.57 & 58.10 & 56.25 \\
DPF-Net~\cite{klokov2020dpfnet} & 75.18 & 65.55 & 62.00 & 58.53 & 62.35 & 54.48 \\
SoftFlow~\cite{kim2020softflow} & 76.05 & 65.80 & 59.21 & 60.05 & 64.77 & 60.09 \\
SetVAE~\cite{kim2021setvae} & 76.54 & 67.65 & 58.84 & 60.57 & 59.94 & 59.94 \\
DPM~\cite{luo2021dpm} & 76.42 & 86.91 & 60.05 & 74.77 & 68.89 & 79.97 \\
PVD~\cite{zhou2021pvd} & 73.82 & 64.81 & 56.26 & 53.32 & 54.55 & 53.83 \\
LION~\cite{zeng2022lion} & 67.41 & 61.23 & 53.70 & 52.34 & 53.32 & 54.55 \\
DiT-3D~\cite{mo2023dit} & 69.42 & 65.08 & 55.59 & 54.91 & 53.87 & 53.02 \\
MeshDiffusion~\cite{liu2023meshdiffusion} & 66.44 & 76.26 & 53.69 & 57.63 & 81.43 & 87.84 \\
MLPCM~\cite{du2024mlpcm} & 65.12 & 58.70 & 51.48 & 50.06 & 51.17 & 48.92 \\
NSOT~\cite{hui2025not}  & 68.64 & 61.85 & 55.51 & 57.63 & 59.66 & 53.55 \\
3DQD~\cite{li2023generalized} & \textbf{56.29} & \textbf{54.78} & 55.61 & 52.94 & 55.75 & 52.80 \\
PVD-DDIM~\cite{song2020denoising} & 76.21 & 69.84 & 61.54 & 57.73 & 60.95 & 59.35 \\
PSF~\cite{wu2023fast} & 71.11 & 61.09 & 58.92 & 54.45 & 57.19 & 56.07 \\
ShapeGF~\cite{cai2020shapegf} & 80.00 & 76.17 & 68.96 & 65.48 & 63.20 & 56.53 \\
\midrule
ConTiCoM (S=1) & 70.20 & 66.56 & 55.67 & 52.14 & 53.89 & 51.97 \\
ConTiCoM (S=2) & 64.89 & 61.77 & \textbf{54.30} & \textbf{50.52} & \textbf{53.30} & \textbf{51.40} \\
\bottomrule
\end{tabular}%
\label{tab:sing_gen}
\end{table}

\begin{table*}[htb]
    \centering
    \caption{Single-class generation results on ShapeNet dataset from PointFlow~\cite{yang2019pointflow}.We use 1-NNA$\downarrow$ as generation metric for evaluation. Training and test data normalized individually into $[-1, 1]$.}
    \label{tab:single_class_individuell}
    \begin{tabular}{lllllll}
        \hline
        \multirow{2}{*}{\textbf{Method}} & 
        \multicolumn{2}{c}{\textbf{Airplane}} & 
        \multicolumn{2}{c}{\textbf{Chair}} & 
        \multicolumn{2}{c}{\textbf{Car}} \\
        %\cline{2-7}
        & \textbf{CD} & \textbf{EMD} & 
          \textbf{CD} & \textbf{EMD} & 
          \textbf{CD} & \textbf{EMD} \\
        \hline
        LION~\cite{zeng2022lion} & 76.30 & 67.04 & 56.50 & 53.85 & 59.52 & 49.29 \\
        Tree-GAN~\cite{liu2018treegan} & 97.53 & 99.88 & 88.37 & 96.37 & 89.77 & 94.89 \\  
        SP-GAN~\cite{li2021spgan} & 94.69 & 93.95 & 72.58 & 83.69 & 87.36 & 85.94 \\
        PDGN~\cite{hui2020progressive} & 94.94 & 91.73 & 71.83 & 79.00 & 89.35 & 87.22 \\
        GCA~\cite{zhang2021learning} & 88.15 & 85.93 & 64.27 & 64.50 & 70.45 & 64.20 \\
        ShapeGF~\cite{cai2020shapegf} & 81.23 & 80.86 & 58.01 & 61.25 & 61.79 & 57.24 \\
        MLPCM(TM)~\cite{du2024mlpcm} & \textbf{73.28} & \textbf{63.08} & 56.20 & 53.16 & 58.31 & \textbf{47.74} \\
        MLPCM(LCM)~\cite{du2024mlpcm} & 75.56 & 66.85 & 58.58 & 55.32 & 61.28 & 49.91 \\
        \hline
        ConTiCoM-3D (S=1) & 74.98 & 70.01 & 57.68 & 56.78 & 59.77 & 53.05 \\
        ConTiCoM-3D (S=2) & 74.22 & 68.54 & \textbf{56.11} & \textbf{53.09} & \textbf{55.43} & 54.22 \\
        \hline
    \end{tabular}
\end{table*}

\noindent\textbf{Sampling Time:} We also report the sampling time per shape in seconds. As shown in App.~\ref{app:inference_time}, ConTiCoM-3D enables fast inference at 0.22 seconds per sample, which is substantially faster than diffusion-based models such as PVD (29.9\,s) and DPM (22.8\,s), and competitive with efficient baselines such as PSF (0.04\,s) and MLPCM (0.18\,s). Although not the absolute fastest, our model offers a balance between speed and fidelity, making it well-suited for real-time applications without compromising geometric accuracy.

\begin{figure*}[htb]
  \centering
  \includegraphics[width=0.95\textwidth]{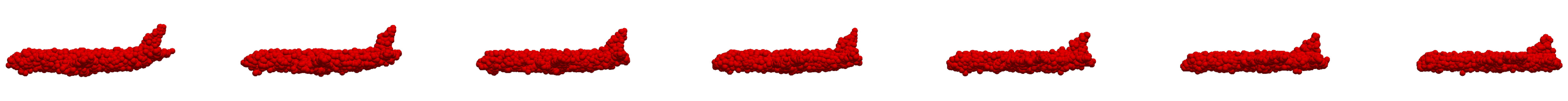}
  \caption{Latent interpolation results with ConTiCoM-3D.}
  %\caption{The top row shows interpolations generated by ConTiCoM-3D, and the bottom row by ConTiCoM-3D. ConTiCoM-3D exhibits smoother transitions and better structural preservation throughout the sequence.}
  \label{fig:interpolation}
\end{figure*}

\noindent\textbf{Interpolation.}  
We probe representation continuity by interpolating between noise vectors $z_1, z_2 \sim \mathcal{N}(0,I)$. Intermediate shapes are generated with $z_\alpha = (1-\alpha)z_1+\alpha z_2$ and $\hat{x}_0 = f_\theta(x_T,T_{\max})$. Figure~\ref{fig:interpolation} shows smooth and structurally consistent transitions, confirming that ConTiCoM-3D learns a continuous geometry-aware flow. 

\subsection{Ablation Studies}
\label{sec:ablation_main}
We perform extensive ablations to evaluate the contribution of each design choice in ConTiCoM-3D. So, we summarize the main findings here. The complete numerical results are provided in Appendix~\ref{sec:additional_results}, where all tables are reported.

\paragraph{Loss components.}
We disentangle the effect of the analytic FM loss and the Chamfer reconstruction loss. As shown in Appendix~\ref{sec:additional_results}, FM-only training produces diverse but geometrically imprecise shapes, while Chamfer-only training improves fidelity at the cost of mode collapse. The combination of FM and Chamfer with adaptive weighting achieves the best balance across 1-NNA, MMD, COV, and JSD metrics.

\paragraph{Noise schedule.}
We compare TrigFlow with linear and cosine schedules. The results (Appendix~\ref{sec:additional_results}) confirm that TrigFlow provides superior stability and fidelity for few-step generation, validating our choice of forward process.

\paragraph{Sampling steps.}
We examine the inference quality for $S=1,2,4$ with both Euler and Heun solvers. As detailed in Appendix~\ref{sec:additional_results}, performance improves significantly from one to two steps, with diminishing returns beyond. Heun provides slight gains for single-step sampling at a modest computational cost.

\subsection{Multi-Class 3D Shape Generation}
\label{sec:multiclass_gen}
Next, we evaluate unconditional multi-class generation on ShapeNet-vol, where a single model is trained jointly across 13 categories. This setting is more challenging due to the multimodal nature of the shape distribution.  

\begin{table*}[htb]
\centering
\caption{Generation metrics (1-NNA$\downarrow$) on 13 classes of ShapeNet-vol.}
    \label{tab:multi_class}
    \begin{tabular}{lcc}
        \hline
        \textbf{Method} & \textbf{CD} & \textbf{EMD} \\
        \hline
        Tree-GAN~\cite{liu2018treegan} & 96.80 & 96.60 \\
        PointFlow~\cite{yang2019pointflow} & 63.25 & 66.05 \\
        ShapeGF~\cite{cai2020shapegf} & 55.65 & 59.00 \\
        SetVAE~\cite{kim2021setvae} & 79.25 & 95.25 \\
        PDGN~\cite{hui2020pdgn} & 71.05 & 86.00 \\
        DPF-Net~\cite{klokov2020dpfnet} & 67.10 & 64.75 \\
        DPM~\cite{luo2021dpm} & 62.30 & 86.50 \\
        PVD~\cite{zhou2021pvd} & 58.65 & 57.85 \\
        LION~\cite{zeng2022lion} & 51.85 & 48.95 \\
        MLPCM(TM)~\cite{du2024mlpcm} & 50.17 & 47.84 \\
        MLPCM(LCM)~\cite{du2024mlpcm} & 53.85 & 52.45 \\
        \hline
        ConTiCoM-3D (S=1) & 49.30 & 46.50 \\
        ConTiCoM-3D (S=2) & \textbf{48.90} & \textbf{45.21} \\
        \hline
    \end{tabular}
\end{table*}

\noindent\textbf{Results:} Table~\ref{tab:multi_class} shows that ConTiCoM-3D achieves the lowest 1-NNA scores in both CD and EMD, outperforming strong baselines including LION~\cite{zeng2022lion} and MLPCM~\cite{du2024mlpcm}. The variant ConTiCoM-3D maintains strong fidelity while enabling real-time sampling, exceeding other efficient models. By avoiding latent encodings and modeling consistency directly in raw 3D space, our method delivers coherent, high-fidelity samples without iterative refinement or encoder bottlenecks. The results demonstrate that ConTiCoM-3D generalizes effectively from class-specific to multi-class settings within a unified, scalable framework.

\section{Discussion and Conclusion}
\label{sec:conclusion}
We presented \textbf{ConTiCoM-3D}, the first continuous-time consistency model for 3D point cloud generation that operates directly in raw point space. Unlike prior approaches that rely on diffusion teachers, Jacobian–vector products, or latent compression, ConTiCoM-3D combines an analytic flow-matching objective with a lightweight Chamfer reconstruction loss, producing stable training and efficient one- to two-step inference. This design preserves geometric fidelity while avoiding the bottlenecks and instabilities of previous paradigms.

Experiments on ShapeNet benchmarks demonstrate that ConTiCoM-3D achieves state-of-the-art performance among teacher-free methods. In single-class settings, our model outperforms GAN- and VAE-based baselines and surpasses recent diffusion and latent consistency models in both Chamfer and Earth Mover’s metrics, while requiring only two evaluations at inference. In the more challenging multi-class generation task, ConTiCoM-3D generalizes effectively across 13 categories, achieving the lowest 1-NNA scores under both CD and EMD. Qualitative results further show smooth interpolations and coherent shape transitions, confirming that our geometry-aware training encourages continuous and robust generative flows.

The ablation studies highlight three key findings. First, Chamfer supervision is essential to improve spatial fidelity without sacrificing sample diversity. Second, the TrigFlow schedule is critical for stable few-step generation, outperforming linear and cosine baselines. Third, most quality improvements are saturated at two sampling steps, confirming the efficiency of our design. Together, these results validate that our contributions are necessary and sufficient for efficient high-quality 3D generation.

ConTiCoM-3D offers a scalable and practical solution for fast 3D point cloud synthesis, making it well suited for interactive robotics, AR/VR, and medical applications where both speed and accuracy are critical. Future work may extend our framework toward conditional and guided generation, scene-level modeling, and handling high-resolution or partial scans. Another promising direction is to explore hybrid training objectives that integrate consistency with diffusion-style guidance, potentially improving controllability and robustness in more complex 3D settings.

\section*{Acknowledgements}
This research as part of the projects LaiLa and EKI is funded by dtec.bw – Digitalization and Technology Research Center of the Bundeswehr, which we gratefully acknowledge. dtec.bw is funded by the European Union – NextGenerationEU. For computing we used the ISCC system. The cloud solution ISCC have been provided by the project hpc.bw, funded by dtec.bw -- Digitalization and Technology Research Center of the Bundeswehr. dtec.bw is funded by the European Union - NextGenerationEU.

\bibliographystyle{plain}
\bibliography{references}

%%%%%%%%%%%%%%%%%%%%%%%%%%%%%%%%%%%%%%%%%%%%%%%%%%

\clearpage
\appendix

\section{Inferencing}
\label{app:inference_time}

As shown in Table~\ref{tab:table_inferencing}, our ConTiCoM-3D does not achieve the lowest runtimes reported by purely distilled one-step models such as PSF or MLPCM. However, these approaches often sacrifice geometric fidelity for speed, as reflected in their higher Chamfer Distance (CD) and Earth Mover’s Distance (EMD) scores~\cite{fan2017point,yang2019pointflow} (see tables in Section~\ref{sec:experiments}). In contrast, ConTiCoM-3D consistently generates shapes within 0.2-0.4 seconds using only 1–4 steps, while preserving competitive quality. This balance is critical: rather than pushing efficiency at the expense of fidelity, our method demonstrates that near real-time inference can be achieved without degradation of shape detail. We therefore position ConTiCoM-3D not as the absolute fastest option but as a principled design that sustains quality under strict runtime constraints, which we regard as more relevant and meaningful for fair cross-method comparison.

\begin{table}[h]
\centering
\caption{Inference efficiency comparison across methods. We report inference steps and runtime}
\label{tab:table_inferencing}
\begin{tabular}{lcc}
\hline
Model / Method              & Inference Steps & Time (s) \\
\hline
LION (DDPM)~\cite{zeng2022lion}            & 1000            & 27.09    \\
LION (DDIM)~\cite{zeng2022lion}            & 1000            & 27.09    \\
LION (DDIM)~\cite{zeng2022lion}            & 100             & 3.07     \\
LION (DDIM)~\cite{zeng2022lion}            & 10              & 0.47     \\
DPM~\cite{luo2021dpm}                      & 1000+           & 22.80    \\
PVD~\cite{zhou2021pvd}                     & 1000            & 29.90    \\
PVD-DDIM~\cite{song2020denoising}          & 100             & 3.15     \\
1-GAN~\cite{achlioptas2018learning}        & 1               & \textbf{0.03}     \\
SetVAE~\cite{kim2021setvae}                & 1               & \textbf{0.03}     \\
PSF~\cite{wu2023fast}                      & 1               & 0.04     \\
MLPCM (Du et al.)~\cite{du2024mlpcm}       & 1   & $<$0.5     \\
\hline
\multirow{3}{*}{ConTiCoM-3D (ours)}        & 1               & 0.22     \\
                                          & 2               & 0.37     \\
                                          & 4               & 0.42     \\
\hline
\end{tabular}
\end{table}

\section{Training}
\label{sec:architecture}

\subsection{Architecture}
We instantiate $F_\theta$ with a PVCNN-style UNet. 
The network consumes a point set $x_t\in\mathbb{R}^{B\times 3\times N}$ and a scalar timestamp $t$, and predicts a velocity field 
$F_\theta(x_t/\sigma_d,t)\in\mathbb{R}^{B\times 3\times N}$. 
Time is injected through a sinusoidal embedding (such as in positional encodings~\cite{vaswani2017attention}) followed by a two-layer MLP, which is added to the global features of the bottleneck.

The encoder consists of four stages, each implemented as a \emph{PVConv} block that fuses voxelized 3D convolutional features with pointwise MLP features, following PVCNN~\cite{liu2019pvcnn}. 
The channel schedule follows 
\[
\text{dims} = [64,\,128,\,256,\,512,\,1024],
\]  
corresponding to \texttt{dim\_mults}=[2,4,8,16]. 
At stage $\ell$, the voxel resolution is set to $r_\ell = 8 \cdot 2^{\ell-1}$. 
The bottleneck aggregates features with global max pooling, adds the time embedding, and refines the representation with two SharedMLP+LayerNorm blocks.

The decoder mirrors the encoder through four \emph{PointNet++ Feature Propagation (FP)} modules~\cite{qi2017pointnet++}. 
Each FP block interpolates features from a coarser resolution back to a finer one, concatenates them with the corresponding encoder skip features, and processes the result with a SharedMLP. 
Finally, a SharedMLP head outputs three channels, which correspond to the predicted velocity vectors.

The complete layer structure is summarized in Table~\ref{tab:arch}, which reflects the implementation with \texttt{dim\_mults}=[2,4,8,16].

\subsection{Hyperparameter Settings}
We follow the TrigFlow parameterization~\cite{lu2024simplifying}, also given in the main paper (Eq.~\ref{eq:trigflow}), with $\sigma_d=1.0$ and $T_{\max}=\tfrac{\pi}{2}$:
\[
x_t=\cos t\,x_0+\sin t\,z,\quad z\sim\mathcal{N}(0,\sigma_d^2 I).
\]

\begin{table*}[htb]
\centering
\small
\setlength{\tabcolsep}{6pt}
\begin{tabular}{l l l l}
\toprule
Stage & Operation & Channels (in$\rightarrow$out) & Notes \\
\midrule
Input & -- & $3$ & Point coordinates \\
Time  & Sinusoidal + MLP & $64\rightarrow 1024$ & Added at bottleneck \\
\midrule
Enc-1 & PVConv($k=3$, res=8)   & $3\rightarrow 128$   & $C_1=64\cdot 2$ \\
Enc-2 & PVConv($k=3$, res=16)  & $128\rightarrow 256$ & $C_2=64\cdot 4$ \\
Enc-3 & PVConv($k=3$, res=32)  & $256\rightarrow 512$ & $C_3=64\cdot 8$ \\
Enc-4 & PVConv($k=3$, res=64)  & $512\rightarrow 1024$ & $C_4=64\cdot 16$ \\
\midrule
Bottleneck & SharedMLP+LN $\times2$ & $1024\rightarrow 1024$ & add time feature \\
\midrule
Dec-4 & FP + skip(Enc-4) & $(1024+1024)\rightarrow 512$ & upsample coords \\
Dec-3 & FP + skip(Enc-3) & $(512+512)\rightarrow 256$ &  \\
Dec-2 & FP + skip(Enc-2) & $(256+256)\rightarrow 128$ &  \\
Dec-1 & FP + skip(Enc-1) & $(128+128)\rightarrow 64$  &  \\
Head  & SharedMLP & $64\rightarrow 3$ & velocity channels \\
\bottomrule
\end{tabular}
\caption{ConTiCoM-3D architecture used for $F_\theta$, instantiated with \texttt{dim\_mults}=[2,4,8,16]. Encoder blocks are PVConv layers with voxel resolutions $[8,16,32,64]$, the bottleneck injects time conditioning, and the decoder consists of FP upsampling modules with skip connections.}
\label{tab:arch}
\end{table*}

\paragraph{Objective.} 
Training is based on single-time supervision combining analytic Flow Matching~\cite{lipman2022flow} (FM) and Chamfer reconstruction~\cite{fan2017point}.  
The FM loss is identical to Eq.~\ref{eq:fm} in the main paper.  
For completeness, we restate the Chamfer reconstruction loss (Eq.~\ref{eq:cd}):
\[
\mathcal{L}_{\mathrm{CD}}=\mathrm{CD}(f_\theta(x_t,t),x_0),
\]
where $\mathrm{CD}$ denotes the \emph{symmetric squared Chamfer distance}, normalized by the number of points.  
The total training objective (Eq.~\ref{eq:total_loss}) is
\[
\mathcal{L}_{\text{total}}=\mathcal{L}_{\mathrm{FM}}+\lambda_{\mathrm{CD}}(t)\,\mathcal{L}_{\mathrm{CD}},
\]
with $\lambda_{\mathrm{CD}}(t)$ an adaptive time-dependent weight, following the ConTiCoM-3D formulation.

\paragraph{Training.} 
We train using Adam~\cite{kingma2014adam} with learning rate $1\times 10^{-4}$ and gradient clipping at $1.0$. 
The batch size is $64$, and the schedule runs for $300$ epochs.
Point clouds are normalized to zero mean and unit radius, with $2048$ points per shape.

\paragraph{Chamfer weighting.} 
Following our approach, we employ a \emph{time-dependent adaptive weighting} $\lambda_{\mathrm{CD}}(t)\in[0.1,0.3]$, which balances FM and Chamfer terms across different noise levels $t$. This adaptive weighting improves stability and fidelity compared to fixed schedules.

\paragraph{Note.}  
As shown in Proposition~\ref{prop:exact_recovery} and proved in Appendix~\ref{sec:proposition_proof}, minimization to zero of the FM loss implies $f_\theta(x_t,t) = x_0$ for all $t \in [0,T_{\max}]$. This means that a perfect velocity regression suffices for exact reconstruction.

\paragraph{Sampling.} 
Single-step generation uses the predictor defined in main paper Eq.~\ref{eq:singlestep}:
\[
\hat{x}_0=f_\theta(x_T,T_{\max}), \quad x_T\sim\mathcal{N}(0,\sigma_d^2 I),
\]
achieving close to real-time inference.  
For optional refinement, we integrate the reverse PF-ODE (Eq.~\ref{eq:pfode_sampler}) using $S$ steps:
\[
x_{t-\Delta}=x_t-\Delta\,\sigma_d\,F_\theta(x_t/\sigma_d,t), \quad \Delta=T_{\max}/S.
\]
By default, we use explicit Euler steps. Optionally, a \emph{Heun proposal}~\cite{lu2024simplifying} can be used for higher-order correction and local error control. Values $S=2$ or $S=4$ already improve fine-scale detail with minimal cost, while larger values of $S$ provide diminishing returns.

\section{Proof of Proposition}
\label{sec:proposition_proof}
We restate Proposition~1 and provide a detailed derivation. We also discuss the role of the flow-matching loss and what happens if it is only approximately satisfied, linking this directly to reconstruction and generation quality.  

\paragraph{Proposition~1 (Closed-Form Recovery).}  
Let the TrigFlow trajectory be given by Eq.~\ref{eq:trigflow}:  
\[
x_t = \cos t \, x_0 + \sin t \, z, 
\qquad t \in [0,T_{\max}], \quad T_{\max}=\tfrac{\pi}{2},
\]
where \(x_0 \in \mathbb{R}^d\) is a data sample and \(z \sim \mathcal{N}(0,\sigma_d^2 I)\).  
Define the predictor as in Eq.~\ref{eq:ctcm_form}:  
\[
f_\theta(x_t,t) = \cos t \, x_t - \sin t \, \sigma_d F_\theta(x_t/\sigma_d,t).
\]
The forward dynamics satisfy Eq.~\ref{eq:forward_trigflow}:  
\[
\frac{dx_t}{dt} = -\sin t \, x_0 + \cos t \, z.
\]
The flow-matching loss is defined in Eq.~\ref{eq:fm}:  
\[
\mathcal{L}_{\mathrm{FM}}
= \big\| \, \sigma_d F_\theta(x_t/\sigma_d,t) - (\cos t \, z - \sin t \, x_0) \,\big\|_2^2.
\]
In the ideal zero-loss case, this implies the analytic regression target
\[
\sigma_d F_\theta(x_t/\sigma_d,t) = \cos t \, z - \sin t \, x_0.
\]
Then, for all \(t\in[0,T_{\max}]\),
\[
f_\theta(x_t,t) = x_0.
\]

\paragraph{Proof.}  
From Eq.~\ref{eq:trigflow},  
\[
x_t = \cos t \, x_0 + \sin t \, z.
\]
Substituting into Eq.~\ref{eq:ctcm_form} gives  
\[
f_\theta(x_t,t) = \cos t \, x_t - \sin t \, \sigma_d F_\theta(x_t/\sigma_d,t).
\]
Using the zero-loss condition implied by Eq.~\ref{eq:fm},  
\[
\sigma_d F_\theta(x_t/\sigma_d,t) = \cos t \, z - \sin t \, x_0,
\]
we obtain
\begin{align*}
f_\theta(x_t,t)
&= \cos t \big( \cos t \, x_0 + \sin t \, z \big)
   - \sin t \big( \cos t \, z - \sin t \, x_0 \big) \\
&= \cos^2 t \, x_0 + \cos t \sin t \, z
   - \sin t \cos t \, z + \sin^2 t \, x_0 \\
&= (\cos^2 t + \sin^2 t) \, x_0 \\
&= x_0.
\end{align*}
Thus, for every \(t \in [0,T_{\max}]\), the predictor exactly recovers the ground-truth data point:
\[
f_\theta(x_t,t) = x_0. \quad \qed
\]

\paragraph{Discussion of Assumptions.}  
The exact recovery critically depends on the flow-matching loss of Eq.~\ref{eq:fm}. This objective enforces
\[
\sigma_d F_\theta(x_t/\sigma_d,t) \approx \cos t \, z - \sin t \, x_0.
\]

- In the **ideal case** (zero loss), the equality holds exactly.

- In **practice**, training and model capacity limitations imply
\[
\sigma_d F_\theta(x_t/\sigma_d,t) = \cos t \, z - \sin t \, x_0 + \varepsilon_t,
\]
where $\varepsilon_t$ denotes the residual regression error.  

Substituting this into Eq.~\ref{eq:ctcm_form} yields
\[
f_\theta(x_t,t) = x_0 - \sin t \, \varepsilon_t.
\]

\paragraph{Consequences of Imperfect Recovery.}

There are three consequences of imperfect recovery: 

\textbf{Error amplification with time.} The deviation from \(x_0\) scales with \(\sin t\). At small \(t\), the error contribution is minimal, so the reconstructions remain close to exact. Near \(T_{\max}=\pi/2\), however, the factor \(\sin t\) is large and inaccuracies become most pronounced.

\textbf{Impact on reconstruction quality.} Imperfect cancellation of the noise \(z\) leads to visible artifacts: reconstructions may appear \emph{blurry, distorted, or biased}. The model fails to perfectly invert Eq.~\ref{eq:trigflow}, so residual noise contaminates the recovered sample.

\textbf{Impact on generation quality.} In generative sampling, these errors manifest as \emph{loss of sharpness}, \emph{inconsistent details}, or \emph{mode bias} where the outputs drift toward regions favored by the imperfect predictor. The neat cancellation between cosine and sine terms, guaranteed by the zero-loss condition of Eq.~\ref{eq:fm}, is thus the mechanism that allows clean recovery. When it fails, quality degradation becomes noticeable.

\iffalse
1. \textbf{Error amplification with time.}  
   The deviation from \(x_0\) scales with \(\sin t\). At small \(t\), the error contribution is minimal, so reconstructions remain close to exact. Near \(T_{\max}=\pi/2\), however, the factor \(\sin t\) is large, and inaccuracies become most pronounced.  

2. \textbf{Impact on reconstruction quality.} 
   Imperfect cancellation of the noise \(z\) leads to visible artifacts: reconstructions may appear \emph{blurry, distorted, or biased}. The model fails to perfectly invert Eq.~\ref{eq:trigflow}, so residual noise contaminates the recovered sample.  

3. \textbf{Impact on generation quality.}  
   In generative sampling, these errors manifest as \emph{loss of sharpness}, \emph{inconsistent details}, or \emph{mode bias} where outputs drift toward regions favored by the imperfect predictor. The neat cancellation between cosine and sine terms, guaranteed by the zero-loss condition of Eq.~\ref{eq:fm}, is thus the mechanism enabling clean recovery. When it fails, quality degradation becomes noticeable.  
\fi

\paragraph{Conclusion.}  
The derivation confirms that Proposition~1 holds rigorously under the ideal assumption implied by Eq.~\ref{eq:fm}, with \(f_\theta\) acting as an exact inverse mapping along the TrigFlow trajectory in Eq.~\ref{eq:trigflow}. When the assumption is only approximately met, Eq.~\ref{eq:ctcm_form} yields systematic deviations amplified by \(\sin t\), which explains the observed degradation in both reconstruction and generation quality. Proposition~1 therefore serves both as a theoretical guarantee and as a diagnostic principle linking analytic flow-matching to practical performance.  

\begin{remark}
Proposition~1 highlights a precise correspondence between the flow-matching loss (Eq.~\ref{eq:fm}) and perfect data recovery: when the loss is minimized to zero, reconstruction is exact; when it is nonzero, the resulting artifacts directly explain the observable drop in reconstruction and generation fidelity (Fig.~\ref{fig:single_gnerated} in the main text for empirical evidence).
\end{remark}

\section{Limitations of Teacher-Based Supervision in Point Cloud Models}
\label{sec:teacher}
A common strategy in consistency models (CMs) is \textbf{teacher–student distillation}, where a target function $f^-_\theta$ (often a stop-gradient or exponential moving average (EMA) of the current model) supervises predictions over time. The canonical distillation loss is introduced in previous CM works~\cite{song2023consistency,jiang2025consistency,lu2024manicm}:
\begin{equation}
\mathcal{L}_{\text{distill}} = \mathbb{E}_{x_t, t} \left[ \left\| f_\theta(x_t, t) - f^-_\theta(x_{t - \Delta t}, t - \Delta t) \right\|^2 \right],
\label{eq:distillation_loss}
\end{equation}
where $x_{t - \Delta t}$ is computed using a known generative path (e.g., a reverse ODE), and $f^-_\theta$ is a stop-gradient copy or an EMA teacher.

Although this has proven effective in image domains~\cite{song2023consistency,lu2024manicm}, the approach fails when extended to 3D point cloud generation for several fundamental reasons:

\paragraph{(1) Absence of reliable perceptual metrics.}
In 2D vision, perceptual metrics such as LPIPS~\cite{zhang2018unreasonable} or CLIP feature distances~\cite{radford2021learning} align well with human similarity judgments. However, point clouds are unordered and sparse representations in $\mathbb{R}^3$ without canonical structure or correspondence. The lack of robust semantic metrics makes it hard to quantify the agreement between teacher and student outputs. Even if both $f_\theta(x_t, t)$ and $f^-_\theta(x_{t - \Delta t}, t - \Delta t)$ represent plausible reconstructions, the Euclidean distance in Eq.~\ref{eq:distillation_loss} may penalize them heavily due to point permutations or geometric shifts. In practice, Chamfer distance~\cite{fan2017point} and Earth Mover’s Distance (EMD)~\cite{rubner2000earth} are the only metrics widely used for point clouds, and even they are limited in semantic sensitivity~\cite{achlioptas2018learning}.

\paragraph{(2) Permutation and alignment mismatch.}
Let $X = \{x_i\}_{i=1}^M$ and $Y = \{y_j\}_{j=1}^M$ be predicted point clouds. Chamfer or EMD distances allow for set-wise comparison, but teacher–student losses like Eq.~\ref{eq:distillation_loss} assume fixed point correspondences. Mathematically, there exists a permutation $\pi$ such that:
\begin{equation}
\min_{\pi \in S_M} \sum_{i=1}^M \left\| f_\theta^{(i)} - f^-_\theta{}^{(\pi(i))} \right\|^2,
\end{equation}
where $S_M$ is the symmetric group. This matching is non-differentiable and thus incompatible with backpropagation. As a result, point-level losses are ill-posed for unordered sets, motivating permutation-invariant architectures such as PointNet and PointNet++~\cite{qi2017pointnet,qi2017pointnet++}.

\paragraph{(3) Inconsistent signal geometry.}
During early training, the teacher $f^-_\theta$ is itself unreliable and may regress to blurred or collapsed outputs. Since $x_t$ and $x_{t - \Delta t}$ come from different noise levels, the corresponding targets $f^-_\theta(x_{t - \Delta t}, t - \Delta t)$ may lie on significantly different manifolds. The student is then penalized for following a different (yet plausible) geometric reconstruction. This causes instabilities and hurts convergence, consistent with reports of teacher collapse in distillation frameworks~\cite{salimans2022progressive,du2024mlpcm,chen2025sana}.

\paragraph{(4) Temporal drift and gradient mismatch.}
The underlying assumption of distillation is temporal smoothness~\cite{song2023consistency,jiang2025consistency}: 
\[
f^-_\theta(x_{t - \Delta t}, t - \Delta t) \approx f^-_\theta(x_t, t) - \Delta t \cdot \partial_t f^-_\theta(x_t, t).
\]
However, in practice, the numerical estimate
\begin{equation}
\frac{f^-_\theta(x_t, t) - f^-_\theta(x_{t - \Delta t}, t - \Delta t)}{\Delta t}
\end{equation}
does not approximate $\partial_t f^-_\theta$ accurately for point clouds due to high variance and poor alignment~\cite{boffi2025build}. Gradient mismatch leads to unstable training and collapse.

\paragraph{(5) Increased memory and runtime.}
Distillation requires caching teacher predictions and computing ODE steps between time pairs. This is particularly prohibitive for large 3D models and long training horizons, where memory and runtime grow linearly with $\Delta t$ supervision steps~\cite{song2023consistency,jiang2025consistency,du2024mlpcm}.

\paragraph{(6) Empirical failure in 3D.}
Previous 3D works report degraded geometric quality and instability when using teacher–student CMs~\cite{du2024mlpcm,chen2025sana}. Our own ablation studies confirm this: teacher-based losses fail to capture fine-grained geometry, leading to blurred, collapsed, or misaligned reconstructions. Quantitatively, they underperform on both Chamfer and Earth Mover’s metrics (see Section~\ref{sec:experiments}).

\paragraph{Conclusion.}
Due to the absence of semantic alignment, point permutation invariance, temporal instability, and computational overhead, teacher–student consistency is \textit{incompatible} with point cloud generation. This motivates our \textbf{teacher-free design}, which replaces teacher-based supervision with: 
(a) a flow-matching regression to the known analytic TrigFlow direction (Eq.~\ref{eq:forward_trigflow}), and  
(b) a permutation-invariant Chamfer loss (Eq.~\ref{eq:cd}),  
yielding a stable, efficient, and geometry-aware training objective.
\newpage
\section{Additional Experimental Results}
\label{sec:additional_results}

\subsection{Ablation Results}
\begin{table*}[t]
\centering
\caption{Loss ablation on ShapeNet~\cite{chang2015shapenet} \textit{airplane}. Metrics: $\downarrow$ lower is better, $\uparrow$ higher is better.}
\label{tab:abl_loss}
\begin{tabular}{lrrrrr}
\toprule
\textbf{Variant} & 1-NNA (CD)$\downarrow$ & 1-NNA (EMD)$\downarrow$ & MMD (CD)$\downarrow$ & COV$\uparrow$ & JSD$\downarrow$ \\
\midrule
FM only & 72.8 & 70.9 & 2.51 & 42.3 & 0.073 \\
Chamfer only & 83.5 & 62.0 & 3.84 & 19.7 & 0.109 \\
FM + Chamfer ($\lambda=0.3$) & \textbf{69.9} & \textbf{64.8} & \textbf{2.18} & \textbf{51.6} & \textbf{0.061} \\
\bottomrule
\end{tabular}
\end{table*}

\begin{table}[htb]
\centering
\caption{Noise schedule ablation (ShapeNet~\cite{chang2015shapenet} \textit{chair}, $S=2$).}
\label{tab:abl_schedule}
\begin{tabular}{lrr}
\toprule
\textbf{Schedule} & 1-NNA (CD)$\downarrow$ & MMD (EMD)$\downarrow$ \\
\midrule
Linear FM & 58.2 & 2.73 \\
Cosine-DDPM & 57.9 & 2.65 \\
TrigFlow (ours) & \textbf{54.3} & \textbf{2.42} \\
\bottomrule
\end{tabular}
\end{table}

\begin{table*}[b]
\centering
\caption{Effect of sampling steps ($S$) on ShapeNet~\cite{chang2015shapenet} \textit{car}.}
\label{tab:abl_steps}
\begin{tabular}{lrrrr}
\toprule
\textbf{Variant} & 1-NNA (CD)$\downarrow$ & EMD$\downarrow$ & Time (s)$\downarrow$ & Memory (GB)$\downarrow$ \\
\midrule
$S=1$ (Euler) & 53.9 & 51.9 & \textbf{0.22} & \textbf{5.1} \\
$S=2$ (Euler) & \textbf{53.3} & \textbf{51.4} & 0.41 & 5.2 \\
$S=4$ (Euler) & 53.0 & 51.1 & 0.79 & 5.3 \\
$S=1$ (Heun) & 53.5 & 51.7 & 0.25 & 5.2 \\
\bottomrule
\end{tabular}
\end{table*}

We report extended ablation experiments to complement the discussion in Sec.~\ref{sec:ablation_main} of the main paper. These results further disentangle the contributions of our loss formulation, noise schedule, and sampling strategy.

\paragraph{Loss components.}
Table~\ref{tab:abl_loss} compares FM, Chamfer reconstruction, and their combination.  
FM only training produces diverse but geometrically imprecise outputs, as indicated by higher EMD and Jensen–Shannon Divergence~\cite{achlioptas2018learning} (JSD) values.  
Chamfer-only training improves fidelity but collapses to limited modes, yielding poor coverage~\cite{achlioptas2018learning} (COV).  
The joint FM+Chamfer objective strikes the best balance, achieving the lowest 1-NNA and Minimum Matching Distance~\cite{achlioptas2018learning} (MMD) along with the highest COV, confirming the complementarity of reconstruction and analytic supervision.

\paragraph{Noise schedule.}
Table~\ref{tab:abl_schedule} evaluates different forward noise schedules on ShapeNet~\cite{chang2015shapenet} class: \textit{chair}.  
Both linear and cosine schedules lead to degraded 1-NNA and MMD compared to our proposed TrigFlow.  
TrigFlow provides more stable training and improved geometric fidelity in the few-step regime, validating its use as the default schedule.

\paragraph{Sampling steps.}
Table~\ref{tab:abl_steps} reports the effect of varying the number of inference steps $S$ on ShapeNet~\cite{chang2015shapenet} class: \textit{car}.  
Performance improves markedly when increasing from $S=1$ to $S=2$, with diminishing returns beyond.  
Although $S=4$ achieves only marginal gains, it doubles the runtime.  
Interestingly, Heun’s solver provides slightly better single-step results than Euler at minimal extra cost.  
In general, $S=2$ with Euler offers the best trade-off between accuracy and efficiency.

\paragraph{Summary.}
Overall, the ablation results confirm that Chamfer reconstruction complements analytic FM by improving fidelity without reducing diversity, that TrigFlow is essential for stable one- and two-step sampling, and that most performance gains are already achieved with $S=2$ steps. These findings support the design of ConTiCoM-3D and demonstrate that its contributions are robust and largely orthogonal to backbone size or external teacher supervision.

\newpage
\subsection{Further Qualitative Results}
\begin{figure*}[b!]
  \centering
  \includegraphics[width=0.95\textwidth]{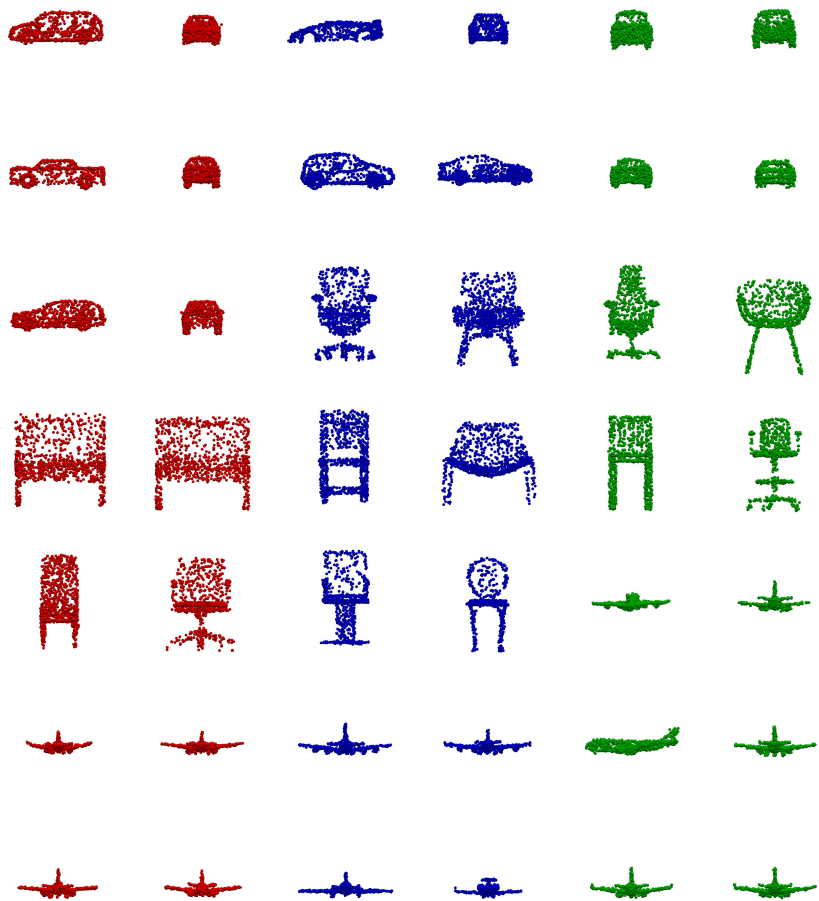}
  \caption{Further qualitative results with ConTiCoM-3D (S=1) on ShapeNet dataset}
    \label{fig:further_single_gnerated}
\end{figure*}

\end{document}